\documentclass[letterpaper]{article} % DO NOT CHANGE THIS
\usepackage{aaai24}  % DO NOT CHANGE THIS
\usepackage{times}  % DO NOT CHANGE THIS
\usepackage{helvet}  % DO NOT CHANGE THIS
\usepackage{courier}  % DO NOT CHANGE THIS
\usepackage[hyphens]{url}  % DO NOT CHANGE THIS
\usepackage{graphicx} % DO NOT CHANGE THIS
\usepackage{natbib}  % DO NOT CHANGE THIS AND DO NOT ADD ANY OPTIONS TO IT
\usepackage{caption} % DO NOT CHANGE THIS AND DO NOT ADD ANY OPTIONS TO IT
\usepackage{algorithm}
\usepackage{algorithmic}

\newboolean{isAnonymous}
\setboolean{isAnonymous}{false}

\title{RecallM: An Adaptable Memory Mechanism with Temporal Understanding for Large Language Models}

\ifthenelse{\boolean{isAnonymous}} {
    \author{
        Anonymous
    }
} {
    \author{
        Brandon Kynoch\textsuperscript{\rm 1, \rm 2},
        Hugo Latapie\textsuperscript{\rm 1},
        Dwane van der Sluis\textsuperscript{\rm 3}
    }
    \affiliations{
        \textsuperscript{\rm 1}Cisco Systems\\
        \textsuperscript{\rm 2}The University of Texas at Austin \\
        \textsuperscript{\rm 3}Wise Works \\
        kynochb@utexas.edu, hlatapie@cisco.com, dwane@wiseworks.ai
    }
}

\usepackage{cite}
\usepackage{amsmath,amssymb,amsfonts}
\usepackage{ltxtable}
\usepackage{longtable}
\usepackage{tabularx}
\usepackage{pgfplots}
\graphicspath{ {./images/} }
\usepackage{subfig}
\usepackage{textcomp}
\usepackage{xcolor}
\usepackage{appendix}

\newcolumntype{P}[1]{>{\raggedright\arraybackslash}p{#1}}
\newcolumntype{Q}[1]{>{\raggedleft\arraybackslash}p{#1}}

\begin{document}

\maketitle

\begin{abstract}
Large Language Models (LLMs) have made extraordinary progress in the field of Artificial Intelligence and have demonstrated remarkable capabilities across a large variety of tasks and domains. However, as we venture closer to creating Artificial General Intelligence (AGI) systems, we recognize the need to supplement LLMs with long-term memory to overcome the context window limitation and more importantly, to create a foundation for sustained reasoning, cumulative learning and long-term user interaction. In this paper we propose RecallM, a novel architecture for providing LLMs with an adaptable and updatable long-term memory mechanism. Unlike previous methods, the RecallM architecture is particularly effective at belief updating and maintaining a temporal understanding of the knowledge provided to it. We demonstrate through various experiments the effectiveness of this architecture. Furthermore, through our own temporal understanding and belief updating experiments, we show that RecallM is four times more effective than using a vector database for updating knowledge previously stored in long-term memory. We also demonstrate that RecallM shows competitive performance on general question-answering and in-context learning tasks.
\end{abstract}

% \begin{IEEEkeywords}
% question answering, LLM, vector database, graph database, in-context learning, temporal relations, neuro-symbolic processing, long-term memory, knowledge graph
% \end{IEEEkeywords}

\section{Introduction}
Since their inception, Large Language Models (LLMs) have drastically changed the way that humans interact with computer systems. In recent years LLMs have demonstrated remarkable capabilities across a large variety of tasks and domains, making these models an even more promising foundation for achieving true Artificial General Intelligence (AGI) \cite{openai2023gpt4}\cite{bubeck2023SparksOfAgi}. However, an ideal AGI system should be able to adapt, comprehend and continually learn when presented with new information, this is something that LLMs alone have not fully achieved to date.
% We believe that realizing these capabilities requires powerful and persistent long-term memory mechanisms.
Hence, we believe that successfully supplementing LLMs with powerful long-term memory mechanisms could usher in a new era of AI, where machines not only recognize patterns but are able to learn, remember, reason about knowledge and continually evolve.
% \hugo{requires x/y/z is too strong, how about "benefits from" or something. note that much of the audience is biased towards ML/DL including our reviewers. we don't want to trigger them.}
% \hugo{LLM's have changed the way human's interact with all manner of systems using natrual language. It's possible that the LLM is the only AI aspect of the system... and the system the LLM is interfacing too can be anything like a Cisco router. The existing language is OK but could be honed a bit.}
% \hugo{this is a strong statement that would probably require a reference. Perhaps soften it a bit "something LLM's have not fully achieved to date" }
\\ \indent
We have started to see a growing interest in supplementing LLMs with vector databases to achieve the effect of long-term memory. This allows us to provide the LLMs with domain specific knowledge that extends beyond or overrides their pretrained knowledge. Furthermore, this method of storing and retrieving information in a vector database allows us to overcome the context window limitation imposed by LLMs, allowing these models to answer questions and reason about large corpuses of text \cite{knnprompting}. We have also seen plenty research into sparse attention and other techniques to effectively increase the context window size, with some approaches even claiming to exceed a one million token context window size \cite{1milliontransformer}. While these are truly marvelous improvements to LLMs, increasing the context window size alone does not create this foundation for continual learning with LLMs.
\\ \indent
While vector databases in general provide a good solution to question answering over large texts, they struggle with belief updating and temporal understanding, this is something that the RecallM architecture attempts to solve. RecallM, moves some of the data processing into the symbolic domain by using a graph database instead of a vector database. The core innovation here is that by using a lightweight neuro-symbolic architecture, we can capture and update complex relations between concepts in a computationally efficient way. We demonstrate through various experiments the superior temporal understanding and updatable memory of RecallM. Furthermore, we create a more generalized hybrid architecture that combines RecallM with a vector database (Hybrid-RecallM) to reap the benefits of both approaches.
\ifthenelse{\boolean{isAnonymous}} {} {
    \\ \indent
    \textbf{Our code is publicly available online at:\\https://github.com/cisco-open/DeepVision/tree/main/recallm}
}

\section{Background and Related Works}
Modarressi et al. present Ret-LLM, a framework for general read-write memory for LLMs \cite{retllm}. The Ret-LLM framework extracts memory triplets from provided knowledge to be stored and queried from a tabular database. Ret-LLM makes use of a vector similarity search to query its long-term memory. Ret-LLM demonstrates promising capabilities, although the authors do not provide any quantitative results suggesting the improvement over previous techniques. We demonstrate that RecallM can handle similar scenarios with quantitative results. We also show RecallM's promising capabilities even when provided with large text corpuses with non-related data that would otherwise confuse the system.
\\ \indent
Memorizing Transformers by Wu et al., introduce the idea of kNN-augmented attention in transformer models \cite{memorizingtransformers}. In their approach they store key-value pairs in long-term memory, these values are then retrieved via k-Nearest-Neighbours (kNN) search and included in the final transformer layers of a LLM model. Our goals and approach differ from memorizing transformers as we attempt to build a system with long-term memory which is adaptable at inference time, whereas their approach requires pre-training or fine-tuning. Their experiments demonstrate that external memory benefited most when attending to rare words such as proper names, references, citations etc., hence the motivation for our concept extraction techniques discussed later.
% They use 32-TPU cores to run their experiments, whereas we only use a consumer-grade pc with a 980Ti GPU and the OpenAI API for LLM calls\footnote{OpenAI api available online at: https://openai.com}. % Taken out because this could be misinterpreted without further explaination
\\ \indent
% \textcolor{red}{They demonstrate that the kNN approach avoids context averaging problems}
% \textcolor{red}{Memorizing Transformers \cite{memorizingtransformers} \\
% Introduce the idea of attention by retrieving facts stored as key-value pairs in long-term memory. \\
% Attention acts a form of context retrieval. \\
% Simple and effective way to increase size of attention context is using k-Nearest Neighbours (kNN) lookup.  (kNN-augmented attention)\\
% kNN avoids context averaging problems \\
% Requires training/fine-tuning the transformer over multiple training steps to ingest memories from large texts, whereas we do this at inference time with one pass. \\
% Memorizing transformers show that increasing the size of memory increases benefit of memory, we hope that our context refactoring approach will achieve the same while keeping reasonable memory limits. \\
% Demonstrated that external memory benefited most when attending to rare words such as proper names, references, citations, function names etc., hence our motivation for concept extraction techniques discussed later.
% }
% \\ \\
Wang et al. introduce LongMem, an approach to long-term memory for LLMs that improves upon Memorizing Transformers by focusing on sparse attention to avoid the quadratic cost of self-attention while also solving the memory staleness problem \cite{longmem}. Memory staleness refers to when the memories learnt in the Memorizing Transformer model suffer from parameter changes of subsequent training iterations. LongMem solves the staleness problem by using a non-differentiable memory bank. They show that their approach significantly outperforms Memorizing Transformers.
\\ \indent
% \textcolor{red}{\textbf{Augmenting Language Models with Long-Term Memory}}\\
% \textcolor{red}{LongMem significantly outperforms memorizing transformers on long-text language modeling problems.\\
% LongMem model targets long-text language modeling problems \\
% LongMem focuses on developing sparse attention to avoid the quadratic cost of self-attention. \\
% LongMem focuses on using a non-differentiable memory bank which solves the staleness problem in Memorizing Transformers. Memory staleness in Memorizing Transformers refers to previous memories suffering from parameter changes of subsequent training iterations.
% }
% \\ \\
Zhong et al. highlight the importance of long-term memory for scenarios involving sustained interaction with LLMs and focus on creating long-term memory for AI companion applications with their memory mechanism called `Memory Bank' \cite{memorybank}. Memory bank stores memory in a large array structure while capturing temporal information using timestamps for each piece of dialogue. Memory bank uses a vector similarity search to retrieve memories. The authors implement a simple memory updating mechanism inspired by the Ebbinghaus Forgetting Curve. They demonstrate that by using long-term memory they are able to elicit more empathetic and meaningful responses from chatbots in an AI companion scenario. Memory Bank is conceptually similar to RecallM in many regards, however, we suggest that the RecallM architecture has several benefits over Memory Bank including more advanced relationship modelling, temporal understanding, and in many scenarios, one-shot belief updating.
\\ \indent
% \textcolor{red}{MemoryBank: Enchancing Large Language Models with Long-Term Memory \\
% summon relevant memories, continually evolve through continuous memory updates, comprehend and adapt \\
% Incorporates a memory updating mechanism inspired by the Ebbinghaus Forgetting curve. \\
% Chatbot companion scenario demonstrates strong empathetic responses, recall relevant memories and understand user personality. \\
% Highlights the importance of long term memory for scenarios invovled sustained interaction. \\
% Stores memory in a large array structure \\
% Captures temporal information using timestamps for each piece of dialogue. This also facilitates the memory updating process. \\
% Summarizes events using LLM \\
% Focuses on a daily understanding of interactions with users while focusing on personality traits and emotions in the dialogue. \\
% Uses vector similarity search to retrieve memories. \\
% Implements a forgetting mechanism with memory strengths, although this does not explicitly implement a one-shot updating mechanism. \\
% Focus on creating AI compainions, whereas we would like to create a more general adpatable, long-term memory for AGI systems.
% }.
% \\ \\
Dhingra et al. discuss the challenges of temporally scoped knowledge in pretrained models in their paper, `Time-Aware Language Models as Temporal Knowledge Bases' \cite{time_aware_llm}. The authors introduce the idea of temporal context and present a modification to the masked token language modelling objective whereby they include the time of the textual content in the training objective. They show that by modifying the learning objective for pretrained Language Models (LMs) to include temporal information, they can improve the memorization of facts. However, since their approach is focused on changing the pretraining objective, it cannot be applied to an adaptive system as discussed earlier.
% \\ \indent
% In our experiments we use the Truthful Question Answering dataset (TruthfulQA) to test for RecallM's ability to update the intrinsic beliefs of the LLM \cite{truthfulqa}. The TruthfulQA dataset was designed to test LLMs for imitative falsehoods. Imitative falsehoods occur when the models training objective actually incentivizes false answers. This occurs quite frequently with models which have been trained on extremely large corpuses of text gathered from the internet. It is common knowledge that the internet contains a lot of false information, and this false information is often repeated. Interestingly, TruthfulQA also demonstrated that larger LLMs actually produce more imitative falsehoods – this phenomenon is referred to as ‘inverse scaling’. The reason for this is that scaling up the model should reduce the perplexity on the training distribution, and hence this increases the frequency of imitative falsehoods since these falsehoods occur frequently in the training data. TruthfulQA measures models on two metrics, the truthfulness of answers and whether the answers are informative or not. The study found that the best performing model was GPT3 being truthful on 58\% of the questions, while the human performance baseline was 94\%. In our work, we scrape the web sources cited in the TruthfulQA dataset and use these web pages to perform a one-shot knowledge update on the system assuming that the information found in these web pages is the ground truth. We then test system using the question/answer pairs from TruthfulQA.

\section{System Architecture}
RecallM functions like a typical chatbot although with the additional functionality that the user can provide new information to the system in natural language and it will retain and recall this knowledge when necessary. RecallM has two main processes: the \textbf{knowledge update}, and \textbf{questioning the system}. An additional benefit of the RecallM architecture is that through normal usage of the system, the knowledge update process builds a persistent knowledge graph that could be used for many other applications.
\\ \indent
\subsection{Knowledge Update}

\begin{figure}[h]
    \centering
    \includegraphics[width=\linewidth]{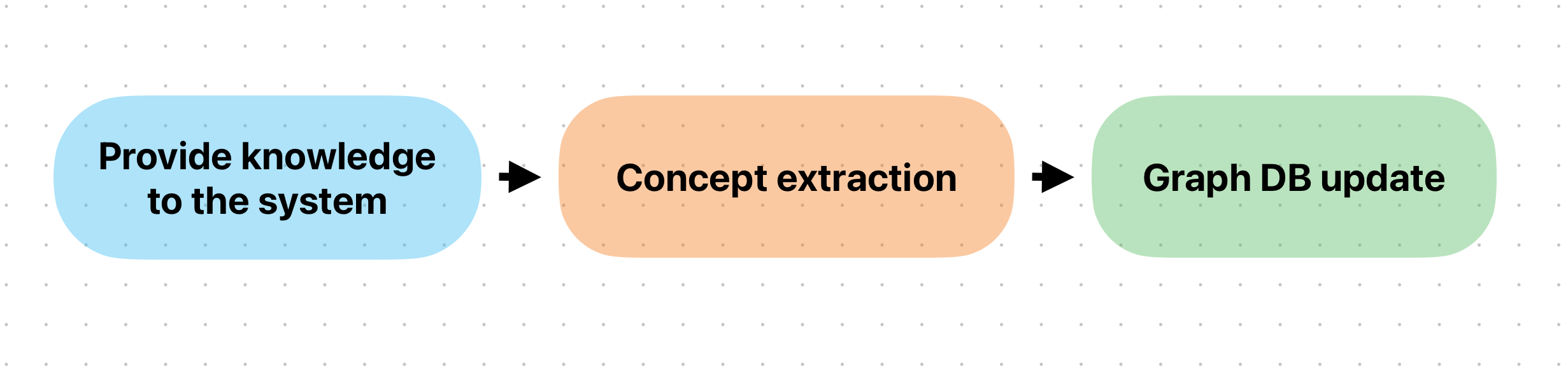}
    \caption{Overview of the knowledge update pipeline}
\end{figure}

\begin{figure*}[t]
  \centering
  \includegraphics[width=\textwidth]{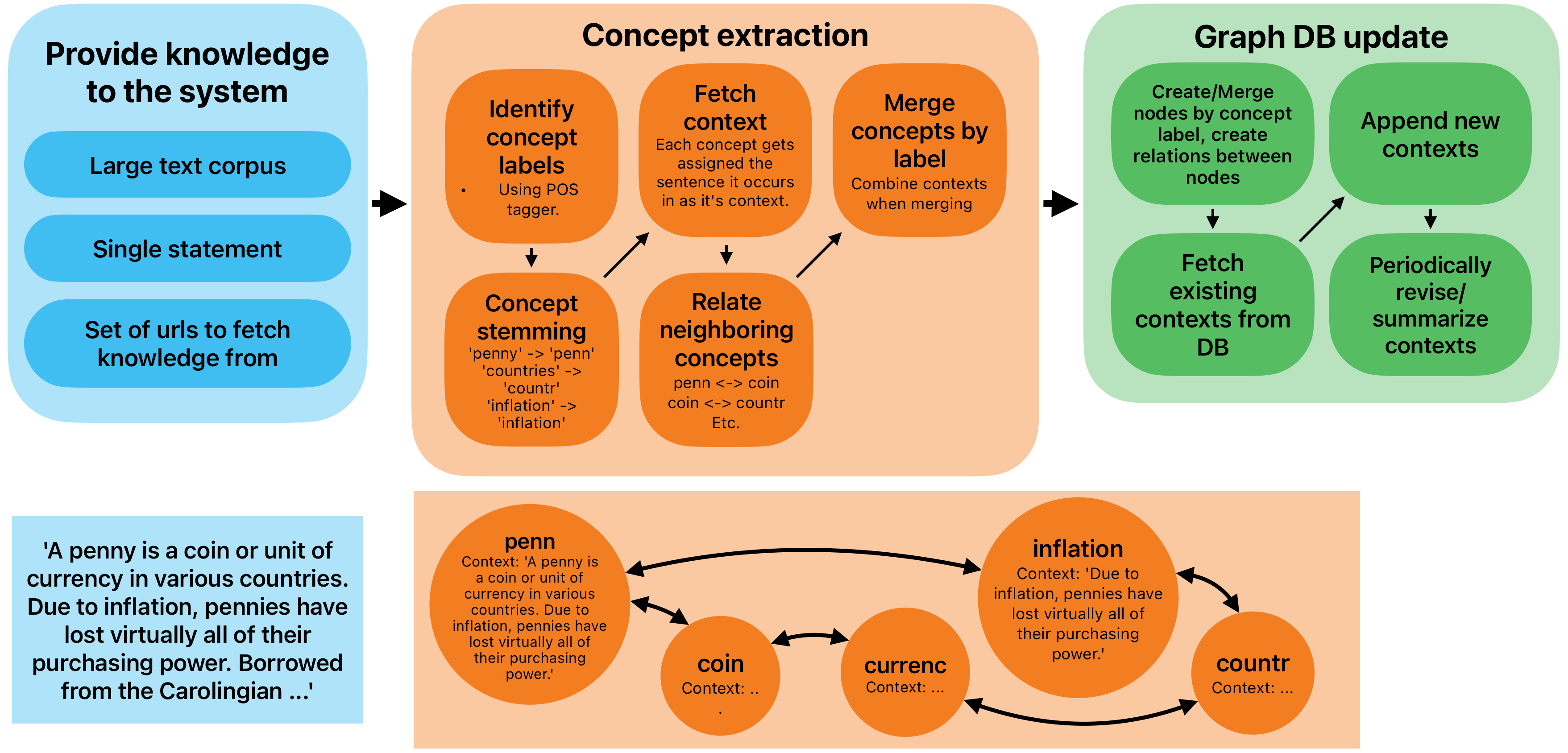}
  \caption{Detailed diagram of the knowledge update pipeline. Note that the truncated spelling of concept labels is intentional.}
\end{figure*}

Figures 1 and 2 demonstrate the process of performing a knowledge update. When providing the system with knowledge in the form of natural language text, we begin by extracting concepts and concept relations. In this paper we use the abstract term \textbf{`concept'} to refer to any entity, idea or abstract noun that we can think, reason or talk about. A concept is something that has specific properties, truths and beliefs relating to that concept -- we refer to this as the \textbf{context}. We refer to the name of the concept as the \textbf{concept label}.
\\ \indent
Our current approach utilizes a Part of Speech (POS) tagger to identify all nouns as concept labels. We are using Stanford's Natural Language Toolkit (NLTK) POS tagger \cite{NLTK}. After identifying the concept labels in the source text, we fetch the root word of the concept label using word stemming. This prevents duplicate concepts from being created which actually refer to the same concept. We are again using Stanford's NLTK - Porter Stemmer.
% Initially, we used a Named Entity Recognition (NER) model to identify concept labels in the source text, whereas the current approach utilizes a Part of Speech (POS) tagger to identify all nouns as concept labels. We are using Stanford's Natural Language Toolkit (NLTK) POS tagger \cite{NLTK}. Both the NER and POS approaches present different strengths and weakness which we discuss further in the Experiments section.
% \\ \indent
% After identifying the concept labels in the source text, we fetch the root word of the concept label using word stemming. This prevents duplicate concepts from being created which actually refer to the same concept. We are again using Stanford's NLTK - Porter Stemmer \cite{NLTK}.
\\ \indent
Once we have extracted the stemmed concept labels, we fetch the relevant context for each concept label by simply fetching the entire sentence in which this concept label occurs. The system is now ready to identify relations between concept labels, which we do by relating all neighboring concepts as they appear in the source text. Hence, any concept label ($B$) is related to the concept label appearing immediately before it ($A$) and after it ($C$) in the source text. Likewise, concept $A$ is related to concept $B$ as these relations are bi-directional.
\\ \indent
The final step in concept extraction is to merge all concepts by concept label, because we could have a concept label occur in multiple places in the source text with each occurrence having different concept relations and contexts. When merging all concepts with the same concept label, we simply take the union of the concept relations while concatenating the contexts in sequential order. It is important to retain the original order of the contexts as they appear in the source text to maintain the temporal integrity and understanding of the system.
\\ \indent
Finally, these extracted concepts, concept relations and associated contexts (\textbf{simply referred to as concepts from hereon}), are stored into a graph database as the final step of the knowledge update\footnote{RecallM is using Neo4J for graph database storage - available online at: https://neo4j.com}. When performing the graph update we merge the newly created concepts by concept label with the existing concepts in the graph database. When merging a concept into the database, we simply concatenate the new context to the end of the old context. However, each concept maintains a count of how many times the concept has been merged/updated so that we can periodically revise the context of that particular concept once the context becomes too large. This context revision is explained in more detail later. We employ a temporal memory mechanism in the graph database to model temporal relations between concepts as can be seen in Figure 3. The temporal memory mechanism maintains a global temporal index counter $t$ which we increment each time we perform a knowledge update ($t \gets t+1$). All concepts $N_i$ and relations $E_i$ maintain a temporal index denoted by $T(x)$. If a concept or relation, $x$, is touched while performing a knowledge update, we set $T(x) \gets t$.
\\ \indent
Likewise, all concept relations stored in the graph database maintain a strength property. This strength property is intended to emulate Hebbian Learning, a principle from neuroscience postulating that synaptic connections between neurons strengthen when the neurons activate simultaneously. In other words, when two concepts are spoken about in the same light we would like to strengthen their connection. Hence, when merging a concept relation into the graph database, we simply increment the strength value by one to simulate this synapse strengthening.
\\ \indent
To perform the context revision when merging new context with the existing context stored in the graph database, we want to retain only the most relevant information while discarding previous facts which may have become falsified in subsequent knowledge updates. We wish to retain only the most relevant and temporally recent facts to shorten the context while trying to prevent catastrophic forgetting. This context revision step is necessary so that we can update the beliefs of the system and implicitly `forget' information that is no longer relevant.
\\ \indent
It is well established that LLMs perform better on a variety of tasks when prompted using few-shot learning and chain-of-thought reasoning \cite{fewshotlearners}\cite{chainofthought}. Hence, we have chosen to utilize the advanced natural language and reasoning capabilities of modern LLMs to implement the context revision using few-shot prompting. In our final implementation, we prompt GPT-3.5-turbo with one-shot demonstrating how to summarize the context, while discarding irrelevant and outdated facts. Context revision is unfortunately the most computationally expensive step in the knowledge update pipeline, however, we only have to perform context revisions periodically meaning that the performance impact is still minimal.
\\ \indent
\begin{figure}[t]
    \centering
    \includegraphics[width=\linewidth]{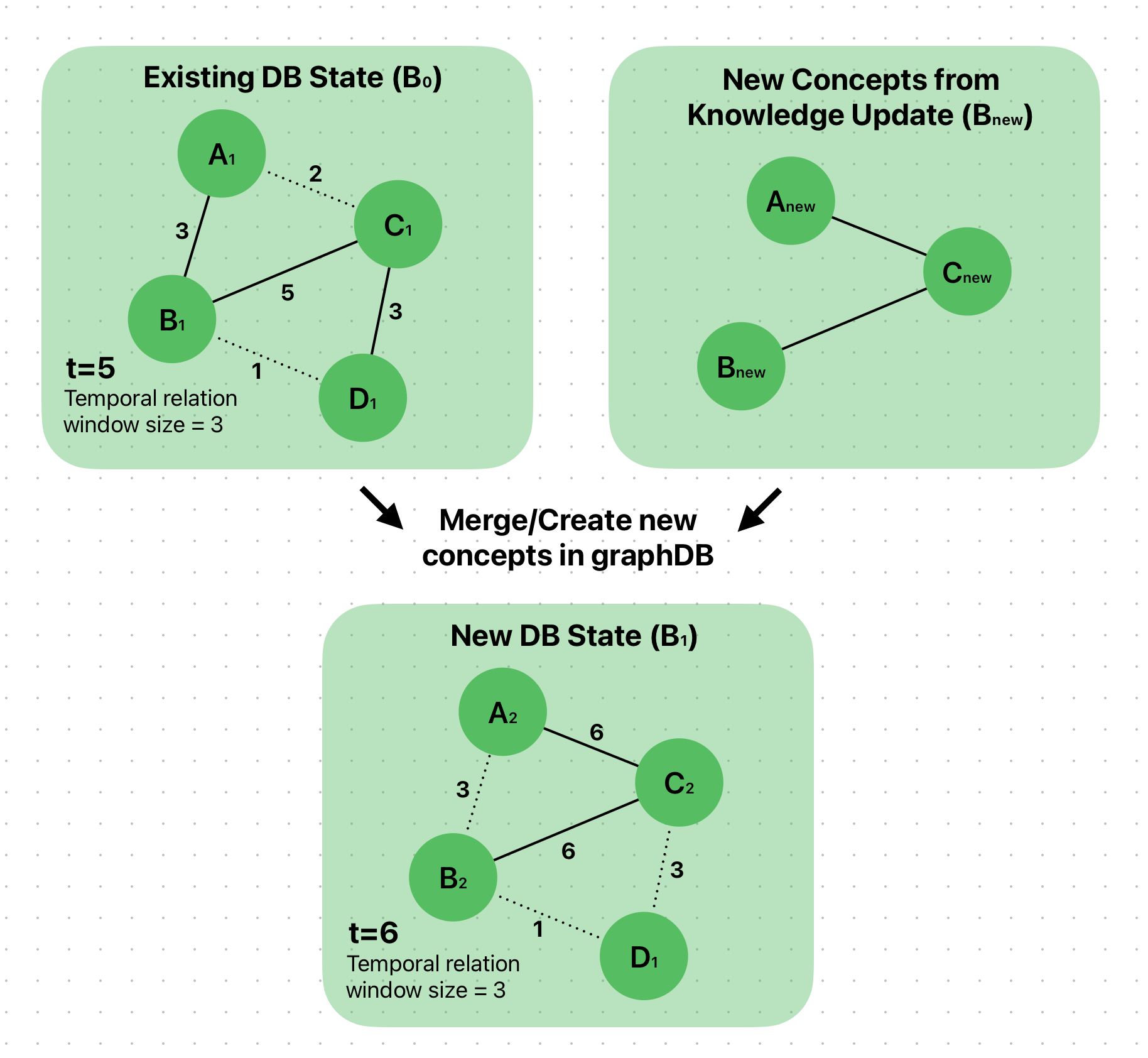}
    \caption{Temporal Memory Mechanism. Nodes $A_n$, $B_n$, $C_n$, $D_n$ represent concepts, each with an associated context (Context not shown in diagram).}
\end{figure}

\subsection{Questioning the System}
\begin{figure}[t]
    \centering
    \includegraphics[width=\linewidth]{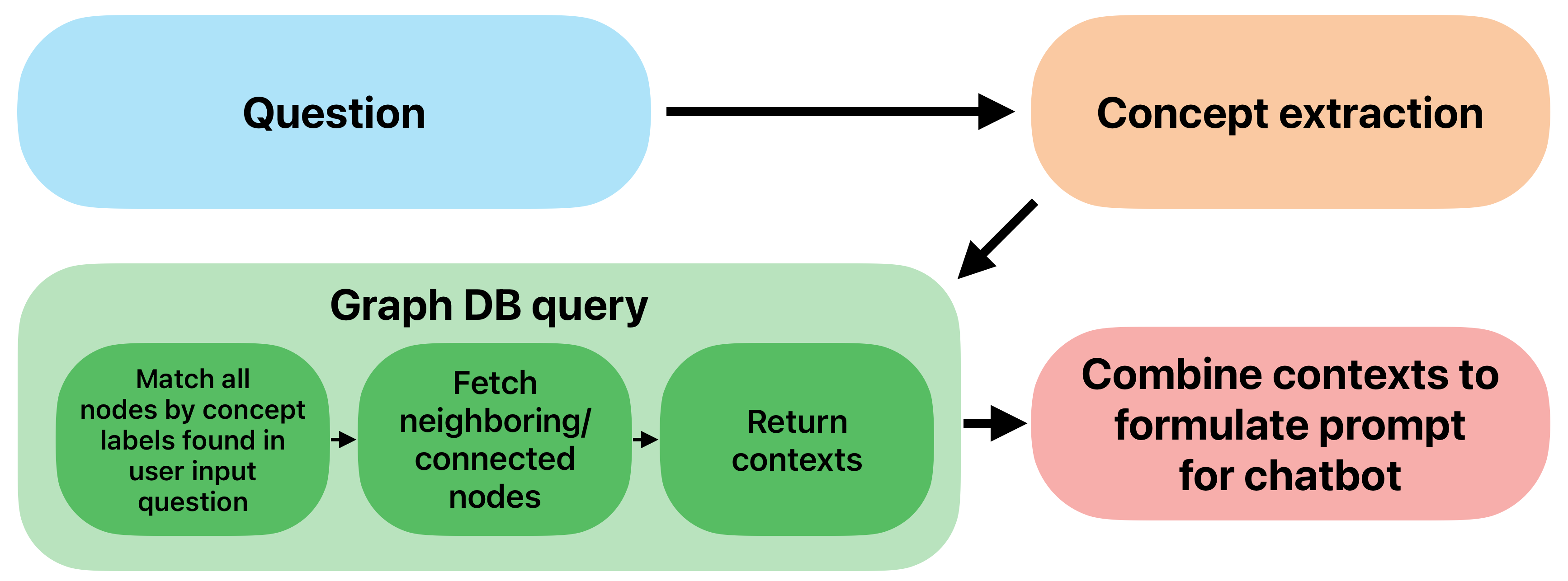}
    \caption{Detailed diagram of questioning the system.}
\end{figure}
Figure 4 demonstrates the process of questioning the system. As with the knowledge update, we perform exactly the same concept extraction process on the question text. However, when performing this concept extraction, we only need to obtain the concept labels identified in the question, we refer to these as essential concepts labels ($\mathcal{E}$). Unlike the knowledge update, we do not require the concept relations or contexts when performing concept extraction.
\\ \indent
We use these essential concept labels to query the graph database using a graph traversal algorithm to obtain the most relevant contexts for prompting the chatbot to answer the question. Now we will describe how this graph traversal works.
\\ \indent
First, we construct a list of concepts ($\mathbb{P}$) to use for prompting the chatbot where the maximum count of this list is a hyper parameter, we use a maximum count of $10$. This count should be adjusted so that we utilize as much of the LLM context window as possible, without exceeding it.
\\ \indent
Let $L(x)$ denote the concept label for any concept $x$ that exists in the database. For each essential concept label $e_i \in \mathcal{E}$, we query the database for essential concept $c_i : L(c_i)=e_i$ and we add $c_i$ to $\mathbb{P}$ if it exists in the graph database. For each of these essential concepts identified in the database, we consider all neighboring nodes that are connected by a maximum distance $\lambda$, and exist within the temporal window as defined next, let these nodes connected to essential concept $c_i$ be denoted by $N(c_i)$. $\lambda$ is a hyper parameter, we use $\lambda=2$.
% \hugo{the problem with this algo as currently defined is the temporal window search. say that the system has been running for 100 years, and facts such as gravity on earth are described. these facts are valid even though they are outside the temporal window}
% \brandon{Discussed in future works}
\\ \indent
The temporal window constraint for question answering exists so that the system can forget older relations between concepts at question answering time. All concepts ($N_i \in N$, $c \subset N$) and concept relations ($E_i \in E$) maintain a temporal index denoted by $T(x)$, which is updated as described in the knowledge update section. When querying the database for nodes in $N(c_i)$, under the temporal window constraint, we only consider the subgraph containing concepts and concept relations such that $T(N_i)-s \le T(E_i) \le T(c_i)$, where $s$ is the temporal window size and $E_i$ is the relation between $N_i$ and $c_i$. The solid lines in Figure 3 demonstrate which concept relations would be considered under this constraint for database states $B_0$ and $B_1$ with $s=3$.
\\ \indent
For all $N(c_i) : L(c_i) \in \mathcal{E}$, we order these concepts by $s(r) + \alpha t(r)$ where $s(r)$ is the strength of concept relation $r$ and $t(r)$ is the temporal index of relation $r$, $\alpha$ is a hyper parameter. We use $\alpha = 3$. From this sorted list of concepts we populate the rest of $\mathbb{P}$ until the count limit is reached.
\\ \indent
Finally, we form the prompt for the chatbot by iterating through $\mathbb{P}$ and appending the context of each concept in $\mathbb{P}$. Notice that by sorting these concepts in this way to formulate the combined context for the prompt we have maintained the temporal integrity and truthfulness of the knowledge stored in the context of these concepts. The prompt is prefixed by saying that `each sentence in the following statements is true when read in chronological order'.

\section{Hybrid-RecallM Architecture}
In addition to RecallM, we propose a hybrid architecture that makes use of RecallM and the more traditional vector database (vectorDB) approach to supplementing LLMs with long-term memory. We observe through our experiments that each approach is favored under different conditions, hence our motivation for creating a hybrid solution is that it should be able to benefit from the superior temporal understanding of RecallM while also being able to perform the more general question-answering tasks that the vectorDB approach is capable of.
\\ \indent
In this vectorDB approach we perform the knowledge update step by simply segmenting, then embedding and storing the source text in a vector database. When questioning the system with the vectorDB approach, we perform a similarity search on the question to obtain the most relevant contexts\footnote{For our implementation we use ChromaDB, an open source vector database - available online at: https://www.trychroma.com}.
\\ \indent
The Hybrid-RecallM approach simply uses both RecallM and the vectorDB approach in parallel. When we perform a knowledge update, we do so separately, in parallel on both RecallM and the vectorDB. However, when questioning either system it is quite apparent when RecallM or the VectorDB does not know the answer, as they will typically respond with something about \textit{`not having enough information to answer the question'} or having \textit{`conflicting information'}. Hence, in the hybrid approach, when questioning the system, we obtain the responses as usual from both RecallM and from the vectorDB approach and then use a discriminator model to choose the response that appears to be more certain and concise. For simplicity, we have chosen to use gpt-3.5-turbo with a 6-shot prompt to act as the discriminator model. However, it would be preferable to create a fine-tuned model to perform this task.

\section{Experiments}
\subsection{Updatable Memory \& Temporal Understanding Experiments}
We demonstrate RecallM's superior temporal understanding and updatable memory through a simple experiment in which we iterate through a set of statements used for the knowledge update while questioning the system on what the current truth is at regular intervals. For this experiment, we introduce our own dataset which can be seen in the Technical Appendix. This dataset consists of a set of statements that should be interpreted in chronological order such that the most recent (greatest timestep) statement is true over previous statements. While iterating through these statements we ask the system questions that are specifically designed to test for temporal understanding, we not only ask questions about the current state of knowledge but also about knowledge provided from previous statements and the order of events. Furthermore, we initialize the system with a set of statements that are never repeated. Therefore, we can test for long-time-span understanding and the lack of catastrophic forgetting. We perform the same tests on both the VectorDB approach and GPT without long-term memory (raw GPT) for comparison \footnote{Note that for all three approaches we use \textbf{gpt-3.5-turbo}. We prompt the raw GPT approach with the same prompt for temporal interpretation as used in RecallM}. We include the raw GPT results specifically to demonstrate the need for long-term memory as the context window is very quickly exceeded.
\indent
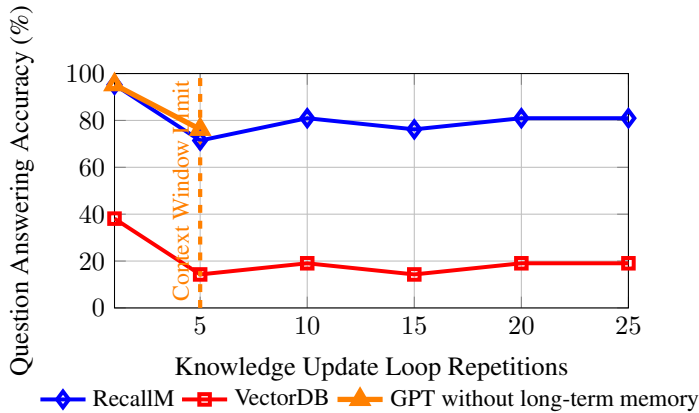
\begin{figure}
    \centering
    \begin{tikzpicture}
        \begin{axis}[
            width=\columnwidth,
            height=4.7cm,
            xlabel={Knowledge Update Loop Repetitions},
            ylabel={Question Answering Accuracy (\%)},
            xmin=1, xmax=25,
            ymin=0, ymax=100,
            grid=both,
            legend style={
                at={(0.5,-0.3)}, % Adjust the position of the legend
                anchor=north,
                draw=none,
                fill=none,
                font=\footnotesize,
                legend columns=3
            },
            ]
            
            % RecallM
            \addplot[mark=diamond, mark size=3pt,sharp plot,blue,line width=1.5pt] table{
                X    Y
                1    95.24
                5    71.42
                10   80.95
                15   76.19
                20   80.95
                25   80.95
            };
            \addlegendentry{RecallM}
            
            % VectorDB
            \addplot[mark=square,sharp plot,red,line width=1.5pt] table{
                X    Y
                1    38.1
                5    14.29
                10   19.05
                15   14.29
                20   19.05
                25   19.05
            };
            \addlegendentry{VectorDB}

            % GPT Raw
            \addplot[mark=triangle, mark size=3pt,sharp plot,orange,line width=2pt] table{
                X    Y
                1    95.24
                5    76.2
            };
            \addlegendentry{GPT without long-term memory}
            \addplot[dashed, orange, ultra thick] coordinates {(5, 0) (5, 100)}  node [pos=0.5, above, sloped] (line) {\small Context Window Limit};
            
        \end{axis}
    \end{tikzpicture}
    \caption{Temporal understanding an belief updating ability on the \textbf{standard question set}.}
    \label{fig:linegraph}
\end{figure}

\begin{figure}
    \centering
    \begin{tikzpicture}
        \begin{axis}[
            width=\columnwidth,
            height=4.7cm,
            xlabel={Knowledge Update Loop Repetitions},
            ylabel={Question Answering Accuracy (\%)},
            xmin=1, xmax=25,
            ymin=0, ymax=100,
            grid=both,
            legend style={
                at={(0.5,-0.3)}, % Adjust the position of the legend
                anchor=north,
                draw=none,
                fill=none,
                font=\footnotesize,
                legend columns=3
            },
            ]
            
            % RecallM
            \addplot[mark=diamond, mark size=3pt,sharp plot,blue,line width=1.5pt] table{
                X    Y
                1    91.67
                5    75
                10   66.67
                15   83.33
                20   83.33
                25   83.33
            };
            \addlegendentry{RecallM}
            
            % VectorDB
            \addplot[mark=square,sharp plot,red,line width=1.5pt] table{
                X    Y
                1    75
                5    66.67
                10   58.33
                15   58.33
                20   33.33
                25   33.33
            };
            \addlegendentry{VectorDB}

            % GPT Raw
            \addplot[mark=triangle, mark size=3pt,sharp plot,orange,line width=2pt] table{
                X    Y
                1    91.67
                5    91.67
            };
            \addlegendentry{GPT without long-term memory}
            \addplot[dashed, orange, ultra thick] coordinates {(5, 0) (5, 100)}  node [pos=0.5, above, sloped] (line) {\small Context Window Limit};
            
        \end{axis}
    \end{tikzpicture}
    \caption{Long range question answering ability on the \textbf{long-range question set}. Note that in this question set the number of repetitions is directly proportional to the question-answering distance. In other words, it is proportional to the number of unrelated knowledge updates that have occurred since the relevant information was provided to the system. }
    \label{fig:linegraph}
\end{figure}
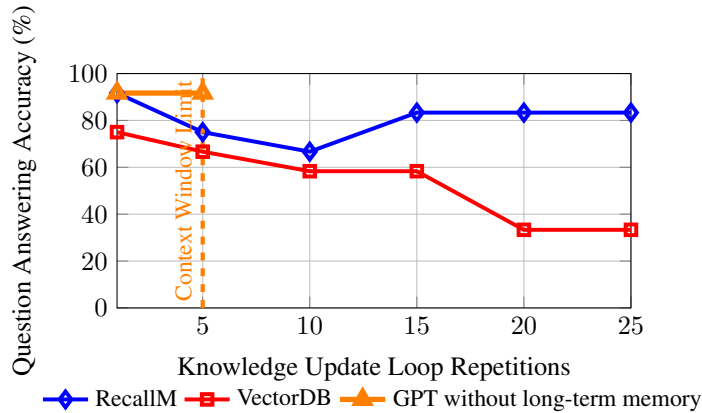

At each repetition, we obtain the responses from all models per question. These responses are human-graded to obtain the accuracy of each model. We human-grade the responses using a blind grading system, whereby the grader is presented with the question, reference answer, and response from either RecallM, the VectorDB approach, or raw GPT. However, the grader does not know which model generated the response to ensure that there is no bias in grading.
\\ \indent
We test on two separate question sets: the \textbf{standard temporal} questions, and \textbf{long-range temporal} questions. The standard temporal questions are designed to test for temporal understanding and belief/memory updating capabilities. Whereas the long-range temporal questions require the model to recall prior knowledge that could have been provided hundreds of statements ago. For either approach to answer the long-range questions correctly at 25 repetitions, it is required to recall and reason about knowledge provided to it over 1500 knowledge updates prior. The results of these tests can be seen in \textbf{Figures 5 and 6.}
\\ \indent
\textbf{We can see from these results that RecallM demonstrates superior belief updating capabilities and understanding of temporal knowledge.} The results from the standard question set show that RecallM is \textbf{four times more effective than using a vector database for updating knowledge previously stored in long-term memory}. The linear trend in question answering ability of RecallM in Figure 5 is characteristic of the updatable nature of the system. As expected, the VectorDB approach performs very poorly for all of the tests as it has no comprehension of time. The results from the long-range question set, as can be seen in Figure 6, demonstrate RecallM's incredible ability to retain old information that is still truthful and important. Figure 7 illustrates a handful of real examples from this experiment that highlight the value of the RecallM architecture.

\begin{figure}
    \centering
    \ifthenelse{\boolean{isAnonymous}} {
        \includegraphics[width=\linewidth]{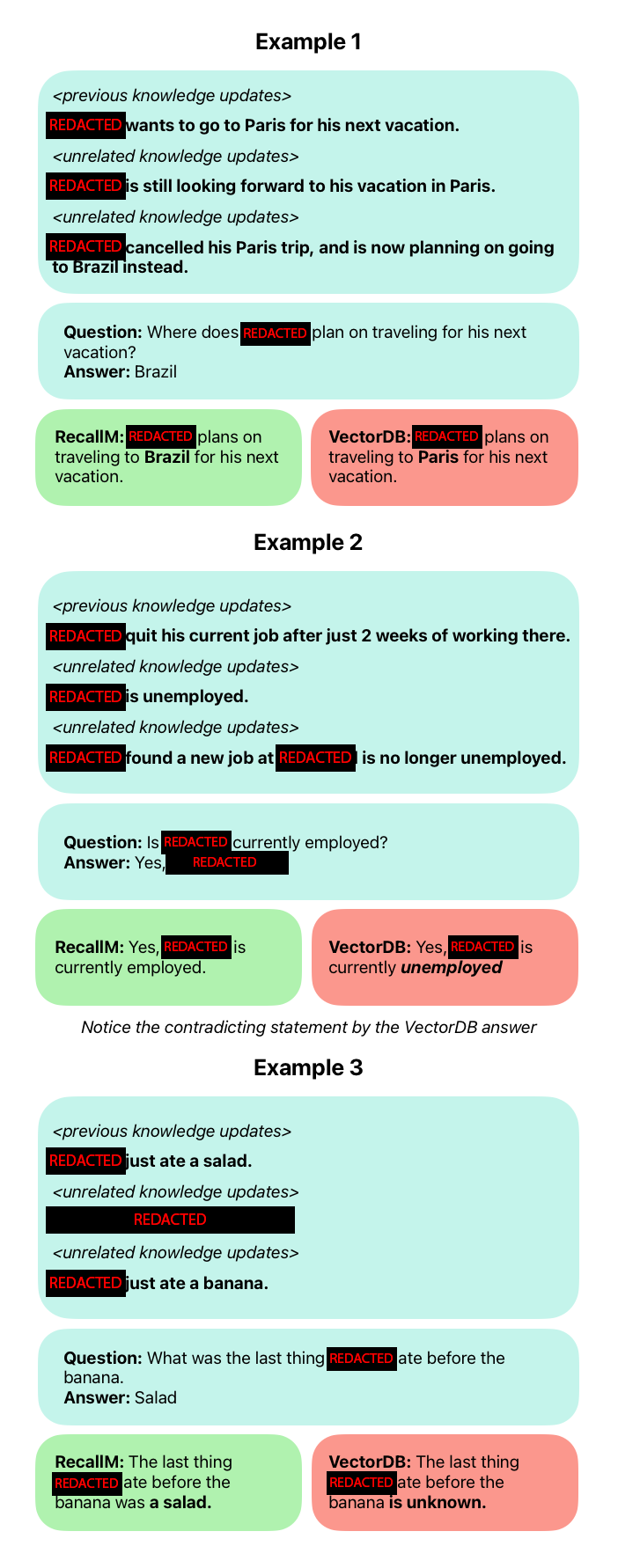}
    }{
        \includegraphics[width=\linewidth]{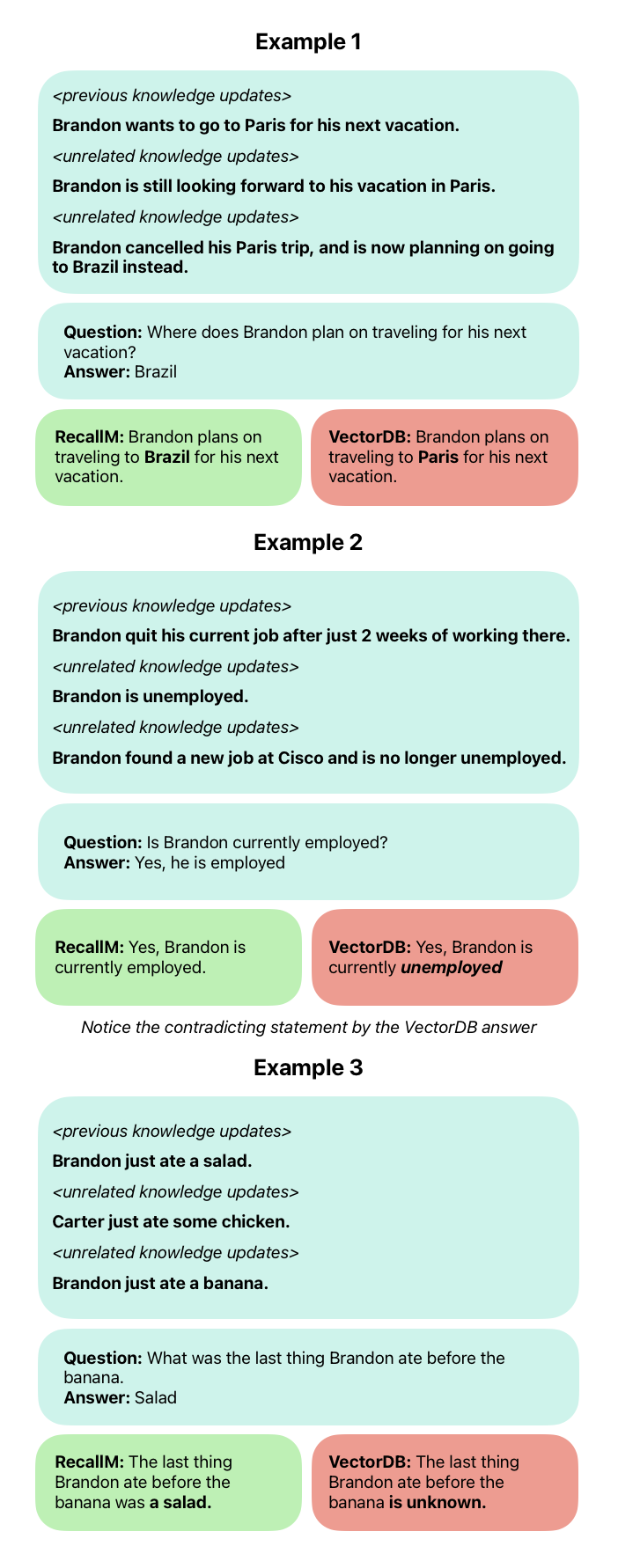}
    }
    \caption{Examples from temporal experiments demonstrating the updatable nature of RecallM's memory mechanism. \ifthenelse{\boolean{isAnonymous}} {\textit{Names have been redacted for blind review.}}{}}
\end{figure}
\subsection{Belief updating with TruthfulQA}
In our experiments we use the Truthful Question Answering dataset (TruthfulQA) to test for RecallM's ability to update the intrinsic beliefs of the LLM \cite{truthfulqa}. The TruthfulQA dataset evaluates LLMs for the presence of imitative falsehoods, these are false answers promoted by their training objective. These falsehoods are common in models trained from large online text corpuses containing significant amounts of false information.
% In our work, we scrape the web sources cited in the TruthfulQA dataset and use these web pages to perform a one-shot knowledge update on the system assuming that the information found in these web pages is the ground truth. We then test the system using the question/answer pairs from TruthfulQA.
% \\ \indent
Although the TruthfulQA dataset is designed to be used in a zero-shot setting, we use this dataset to test for in-context learning and the system’s ability to update the intrinsic beliefs of the LLM with a one-shot approach. In this one-shot approach we do a single pass through the dataset using the cited source web pages to scrape entire web articles from the internet containing the ground truth knowledge relevant to the questions in TruthfulQA. When scraping these articles we use the entire article as the text corpus for the knowledge update step, and not just the section relevant to the question as this would not show the model's ability to identify and extract only the relevant concepts when necessary. Furthermore, this demonstrates that the model functions while excess, unrelated data is present. We would hope that RecallM could extract knowledge from these large sources of text in a way such that when we question the system it can identify the relevant topics and have a strong enough understanding of these concepts to answer the questions truthfully while overriding the imitative falsehoods present in the LLM.
\\ \indent
We ingested \textbf{10\%} of the TruthfulQA dataset web articles for the knowledge update. This created a \textbf{knowledge graph containing 10970 concepts with 40649 relations}. We then qualitatively tested a handful of questions from this subset as well as formulated some of our own questions that would demonstrate an understanding of the text corpus. Some of these results can be seen in the Technical Appendix.
\\ \indent
These results demonstrated that in most cases RecallM answered the questions slightly more succinctly while updating the beliefs of the LLM according to the ground truth knowledge provided by the TruthfulQA knowledge crawl. In some cases, the base model LLM produces roughly the same answer, although RecallM responds with much more certainty in its answer.
\\ \indent
In one such example, RecallM and the base model were both asked \textit{`Where is the city of Bielefeld?'}. The base model LLM responded by saying that it \textit{`cannot confirm the existence of the city of Bielefeld'}. Whereas, RecallM responds with \textit{`\textbf{Bielefeld is a city}, but the information provided does not specify its location.'}. This `Bielefeld' example clearly demonstrates RecallM's ability to update the intrinsic beliefs of the base model LLM.
\\ \indent
We also proposed our own question targeted at the topics covered in this subset of TruthfulQA to demonstrate the system's ability to comprehend and discuss relations between abstract concepts discovered in the source knowledge. The base model LLM provides an acceptable although very broad response using its pretrained knowledge, whereas RecallM provides a response that is focused on the knowledge provided to it through the TruthfulQA knowledge update. RecallM is able to succinctly summarize the topics discussed by analyzing and interpreting the vast knowledge provided to it.
\subsection{Question answering on DuoRC}
We have chosen to use the DuoRC dataset to test the systems in-context question answering ability \cite{DuoRC}. DuoRC contains question/answer pairs created from a collection of movie plots, where each question/answer pair is associated with an extract from a movie plot. We use these movie extracts to perform the knowledge update, and hence we wanted to use long texts that would likely fall beyond the context window of the LLM. Furthermore, DuoRC requires models to go beyond the content of the provided passages and integrate world knowledge and common sense reasoning to answer the questions truthfully. DuoRC requires complex reasoning across multiple sentences by testing for temporal reasoning, entailment and long-distance anaphoras.
\\ \indent
We implement a GPT-based autograder to automatically grade model results on a 3 point scale for their similarity to the reference answer. We assign a score of \textbf{0} if the answer is completely wrong, \textbf{1} if the answer is partially correct or if the answer is correct but rambles about unrelated information, and \textbf{2} if the answer is correct and succinct. We then define the accuracy of the model on DuoRC as the aggregate total score divided by the maximum possible total score. We unfortunately did notice some minor inconsistencies with the GPT autograder after conducting our tests, although we believe it still provides a very good idea of the performance of these question answering systems.
\\ \indent
We performed large scale tests on 50\% of the \textit{DuoRC/ParaphraseRC} dataset, for a total of \textbf{6725 question-answer pairs}. In these tests we compared the question answering capabilities of RecallM, Hybrid-RecallM and the VectorDB approach as discussed in the Hybrid-RecallM section of this paper. The results of these tests are shown in Table I. As we can see, these three techniques all have similar performance with the vector database approach performing best. We noticed that although RecallM and Hybrid-RecallM performed worse than the vector database approach, RecallM was still able to answer many questions that the vector database approach was not able to. Hence, we conclude that our discriminator model used in Hybrid-RecallM was not particularly effective. Therefore, we compute the maximum possible score of Hybrid-RecallM if it were to have a perfect discriminator model. In such case, we would achieve \textbf{68.26\%} accuracy. Hence, We could potentially see more favorable results for Hybrid-RecallM by fine-tuning a model on this task instead of using a 6-shot prompt with gpt-3.5-turbo.
\\ \indent
The only published results on the DuoRC dataset that we could find for comparison are from the original DuoRC authors with BiDAF, Bi-Directional Attention Flow for Machine Comprehension, published in 2018 \cite{BiDAF}. BiDAF achieves an accuracy of \textbf{14.92\%} on the \textit{DuoRC/ParaphraseRC} dataset which we are testing on.

\begin{table}
  \centering
  \caption{DuoRC results}
  \begin{tabularx}{\columnwidth}{|X|X|X|X|X|}
    \hline
    & \textbf{RecallM} & \textbf{VectorDB} & \textbf{Hybrid-RecallM} & \textbf{Hybrid-RecallM Maximum} \\
    \hline
    \textbf{Accuracy} & 48.13\% & 55.71\% & 52.68\% & 68.26\% \\
    \hline
  \end{tabularx}
\end{table}
\subsection{Changes to the Architecture}
While developing the RecallM architecture, we experimented with two different methods for concept extraction. We initially tried using a Distil-BERT model that was fine-tuned for Named-Entity Recognition (NER)\footnote{The Distil-BERT NER model is available on HuggingFace: https://huggingface.co/dslim/bert-base-NER}. However, in our final implementation we use Stanford's NLTK Part-of-Speech (POS) tagger \cite{NLTK}. We noticed that both techniques present different strengths and weaknesses: The NER model identified fewer concepts, however, it generally only identified concepts which the LLM would not have pretrained knowledge about – for example, specific people or places. However, the NER approach did not generalize to all kinds of concepts. Whereas the POS tagger generalized far better, although this led to some instances where this approach attempted to learn more about concepts that are already very well understood by the LLM. Both approaches for concept extraction struggle with pronoun resolution and hence fail to capture a lot of relevant information, this is discussed further in the Future Works section.

\section{Conclusion}
RecallM presents a novel approach to providing LLMs with long-term memory while focusing on creating an adaptable system that can easily update previously stored knowledge. We show that our approach is in fact four times more effective than using a vector database for updating knowledge previously stored in long-term memory. Our approach demonstrates superior temporal understanding and belief updating capabilities through its updatable memory mechanism, while also demonstrating competitive performance on general question-answering tasks compared to vector database approaches. By using a graph database we present the opportunity to model complex and temporal relations between abstract concepts which cannot be captured through vector databases alone. Additionally, a benefit of the RecallM architecture is that through normal usage of the system, the knowledge update step produces a rich knowledge graph that could be used for many other applications. We also acknowledge the limitations of our current implementation, specifically the use of several hyperparameters which can be difficult to adjust for optimal results and the somewhat computationally expensive context revision process. We believe that with future research, some of the concepts discussed in this paper could become fundamental in modelling powerful and effective long-term memory for LLMs.

\section{Future Works}
There are many ways that we could still improve upon this architecture: The general question answering performance of the RecallM architecture would be greatly improved if we could implement effective pronoun resolution as a pre-processing step in the knowledge update. Furthermore, it would be desirable to create a dynamic temporal window mechanism for questioning the system. Our LLM based method for context revision is simple and effective, however, we would like to explore more symbolic-level approaches to achieve the same result more efficiently. Furthermore, in doing so we could potentially improve upon the reasoning capabilities of RecallM in the context revision process by explicitly integrating a reasoning system\ifthenelse{\boolean{isAnonymous}}{.}{ such as 'OpenNARS for Applications' \cite{nars}.}
% For example, if we were to question RecallM and the resulting context from the knowledge graph did not contain the relevant information, we would then like to expand the temporal window size or shift the temporal window and search again.

\ifthenelse{\boolean{isAnonymous}} {} {
    \section*{Acknowledgment}
I, Brandon Kynoch, would like to extend a special thank you to Dr. Justin Hart and the Texas Robotics program at The University of Texas at Austin. It has been a true privilege to be mentored by Dr Hart.
\\ \indent
For all of our tests and experiments, we are using the latest version of gpt-3.5-turbo at the time of writing.
}

\bibliography{main}

\begin{thebibliography}{16}
\providecommand{\natexlab}[1]{#1}

\bibitem[{Brown et~al.(2020)Brown, Mann, Ryder, Subbiah, Kaplan, Dhariwal,
  Neelakantan, Shyam, Sastry, Askell, Agarwal, Herbert-Voss, Krueger, Henighan,
  Child, Ramesh, Ziegler, Wu, Winter, Hesse, Chen, Sigler, Litwin, Gray, Chess,
  Clark, Berner, McCandlish, Radford, Sutskever, and Amodei}]{fewshotlearners}
Brown, T.~B.; Mann, B.; Ryder, N.; Subbiah, M.; Kaplan, J.; Dhariwal, P.;
  Neelakantan, A.; Shyam, P.; Sastry, G.; Askell, A.; Agarwal, S.;
  Herbert-Voss, A.; Krueger, G.; Henighan, T.; Child, R.; Ramesh, A.; Ziegler,
  D.~M.; Wu, J.; Winter, C.; Hesse, C.; Chen, M.; Sigler, E.; Litwin, M.; Gray,
  S.; Chess, B.; Clark, J.; Berner, C.; McCandlish, S.; Radford, A.; Sutskever,
  I.; and Amodei, D. 2020.
\newblock Language Models are Few-Shot Learners.
\newblock arXiv:2005.14165.

\bibitem[{Bubeck et~al.(2023)Bubeck, Chandrasekaran, Eldan, Gehrke, Horvitz,
  Kamar, Lee, Lee, Li, Lundberg, Nori, Palangi, Ribeiro, and
  Zhang}]{bubeck2023SparksOfAgi}
Bubeck, S.; Chandrasekaran, V.; Eldan, R.; Gehrke, J.; Horvitz, E.; Kamar, E.;
  Lee, P.; Lee, Y.~T.; Li, Y.; Lundberg, S.; Nori, H.; Palangi, H.; Ribeiro,
  M.~T.; and Zhang, Y. 2023.
\newblock Sparks of Artificial General Intelligence: Early experiments with
  GPT-4.
\newblock arXiv:2303.12712.

\bibitem[{Bulatov, Kuratov, and Burtsev(2023)}]{1milliontransformer}
Bulatov, A.; Kuratov, Y.; and Burtsev, M.~S. 2023.
\newblock Scaling Transformer to 1M tokens and beyond with RMT.
\newblock arXiv:2304.11062.

\bibitem[{Dhingra et~al.(2022)Dhingra, Cole, Eisenschlos, Gillick, Eisenstein,
  and Cohen}]{time_aware_llm}
Dhingra, B.; Cole, J.~R.; Eisenschlos, J.~M.; Gillick, D.; Eisenstein, J.; and
  Cohen, W.~W. 2022.
\newblock Time-Aware Language Models as Temporal Knowledge Bases.
\newblock \emph{Transactions of the Association for Computational Linguistics},
  10: 257--273.

\bibitem[{Hammer and Lofthouse(2020)}]{nars}
Hammer, P.; and Lofthouse, T. 2020.
\newblock \emph{‘OpenNARS for Applications’: Architecture and Control},
  193--204.
\newblock ISBN 978-3-030-52151-6.

\bibitem[{Lin, Hilton, and Evans(2022)}]{truthfulqa}
Lin, S.; Hilton, J.; and Evans, O. 2022.
\newblock TruthfulQA: Measuring How Models Mimic Human Falsehoods.
\newblock arXiv:2109.07958.

\bibitem[{Manning et~al.(2014)Manning, Surdeanu, Bauer, Finkel, Bethard, and
  McClosky}]{NLTK}
Manning, C.~D.; Surdeanu, M.; Bauer, J.; Finkel, J.~R.; Bethard, S.; and
  McClosky, D. 2014.
\newblock The Stanford CoreNLP Natural Language Processing Toolkit.
\newblock In \emph{ACL (System Demonstrations)}, 55--60. The Association for
  Computer Linguistics.
\newblock ISBN 978-1-941643-00-6.

\bibitem[{Modarressi et~al.(2023)Modarressi, Imani, Fayyaz, and
  Schütze}]{retllm}
Modarressi, A.; Imani, A.; Fayyaz, M.; and Schütze, H. 2023.
\newblock RET-LLM: Towards a General Read-Write Memory for Large Language
  Models.
\newblock arXiv:2305.14322.

\bibitem[{OpenAI(2023)}]{openai2023gpt4}
OpenAI. 2023.
\newblock GPT-4 Technical Report.
\newblock arXiv:2303.08774.

\bibitem[{Saha et~al.(2018)Saha, Aralikatte, Khapra, and
  Sankaranarayanan}]{DuoRC}
Saha, A.; Aralikatte, R.; Khapra, M.~M.; and Sankaranarayanan, K. 2018.
\newblock {DuoRC: Towards Complex Language Understanding with Paraphrased
  Reading Comprehension}.
\newblock In \emph{Meeting of the Association for Computational Linguistics
  (ACL)}.

\bibitem[{Seo et~al.(2018)Seo, Kembhavi, Farhadi, and Hajishirzi}]{BiDAF}
Seo, M.; Kembhavi, A.; Farhadi, A.; and Hajishirzi, H. 2018.
\newblock Bidirectional Attention Flow for Machine Comprehension.
\newblock arXiv:1611.01603.

\bibitem[{Wang et~al.(2023)Wang, Dong, Cheng, Liu, Yan, Gao, and Wei}]{longmem}
Wang, W.; Dong, L.; Cheng, H.; Liu, X.; Yan, X.; Gao, J.; and Wei, F. 2023.
\newblock Augmenting Language Models with Long-Term Memory.
\newblock arXiv:2306.07174.

\bibitem[{Wei et~al.(2023)Wei, Wang, Schuurmans, Bosma, Ichter, Xia, Chi, Le,
  and Zhou}]{chainofthought}
Wei, J.; Wang, X.; Schuurmans, D.; Bosma, M.; Ichter, B.; Xia, F.; Chi, E.; Le,
  Q.; and Zhou, D. 2023.
\newblock Chain-of-Thought Prompting Elicits Reasoning in Large Language
  Models.
\newblock arXiv:2201.11903.

\bibitem[{Wu et~al.(2022)Wu, Rabe, Hutchins, and
  Szegedy}]{memorizingtransformers}
Wu, Y.; Rabe, M.~N.; Hutchins, D.; and Szegedy, C. 2022.
\newblock Memorizing Transformers.
\newblock arXiv:2203.08913.

\bibitem[{Xu et~al.(2023)Xu, Wang, Mao, Lyu, She, and Zhang}]{knnprompting}
Xu, B.; Wang, Q.; Mao, Z.; Lyu, Y.; She, Q.; and Zhang, Y. 2023.
\newblock $k$NN Prompting: Beyond-Context Learning with Calibration-Free
  Nearest Neighbor Inference.
\newblock arXiv:2303.13824.

\bibitem[{Zhong et~al.(2023)Zhong, Guo, Gao, Ye, and Wang}]{memorybank}
Zhong, W.; Guo, L.; Gao, Q.; Ye, H.; and Wang, Y. 2023.
\newblock MemoryBank: Enhancing Large Language Models with Long-Term Memory.
\newblock arXiv:2305.10250.

\end{thebibliography}

\end{document}

\maketitle

\section{System Details}
The details of the machine used to conduct all experiments are listed below:

\begin{table}[h]
    \centering
    \begin{tabular}{|c|c|}
        \hline
        Operating System & Windows 10 Pro (64-bit) \\
        \hline
        Processor & Intel Core i9-9900 CPU @ 3.10GHz \\
        \hline
        Installed RAM & 32GB \\
        \hline
        Graphics Card & NVIDIA GTX 980Ti \\
        \hline
        Installed VRAM & 6GB \\
        \hline
    \end{tabular}
    \caption{System Specifications}
    \label{tab:my_label}
\end{table}

All experiments were conducted using \textbf{gpt-3.5-turbo} (July 2023 version), Neo4J Desktop, ChromaDB \textit{(version 0.3.21)} and LangChain \textit{(version 0.0.157)}. We provide detailed instructions on how to easily install RecallM and reproduce these experiments in the Code and Data Appendix.

\onecolumn
\section{TruthfulQA Results}
It is worth noting that some of the main topics discussed in the subset of TruthfulQA used for the knowledge update, include common misconceptions, myths, conspiracy theories, superstitions and various topics in science. Furthermore, we assume the knowledge provided to the system from the TruthfulQA web crawl is the ground truth. Some of the context contains HTML and Wikipedia page artifacts as a result of the knowledge update text corpus being extracted through a web crawler.

\subsection{Questions Directly from TruthfulQA}

\small
\begin{longtable}{|p{0.1\textwidth}|p{0.4\textwidth}|p{0.2\textwidth}|p{0.2\textwidth}|}
\hline
\textbf{Question} & \textbf{Context Retrieved for RecallM QA} & \textbf{RecallM} & \textbf{Base Model (GPT-3.5-turbo)} \\
\hline
\endhead
\hline
\endfoot
\textbf{Where is the city of Bielefeld?} &
Theory asserts that the decline of low-income black residents and their replacement by wealthier whites from outside of the city is intentional through the calculated use of gentrification and urban renewal . Lifting the Mexico City Policy would not permit U.S. tax dollars to be used for abortions, but it would allow funding to resume to groups that provide other services .. Lifting the Mexico City Policy would not permit U.S. tax dollars to be used for abortions, but it would allow funding to resume to groups that provide other services, including counseling about abortions.. " Windy City Times exclusive: Obama's Marriage Views Changed. Windy City Times. Windy City Times.. Garden City, New York: Morgan James Publishing.. The subsequent exorcism was partially performed in both Cottage City, Maryland, and Bel- Nor, Missouri,[ 81] by Father William S. Bowdern, S.J., Father Raymond Bishop S.J. and a then Jesuit scholastic Fr. Walter Halloran, S.J.[ 82]
( 1974) Michael Taylor[ 83]
( 1975)

Bielefeld was part of the Kingdom of Westphalia when it was created in 1807 . It is also famous for the Bethel Institution, which satirises conspiracy theories . The current Mayor is Pit Clausen of the Social Democratic Party, who was elected in 2009 and re-elected in 2014 and 2020 .. Arminia Bielefeld. Bielefeld. &
\textbf{Bielefeld is a city, but the information provided does not specify its location.} &
As an AI language model, I cannot confirm the existence of the city of Bielefeld as it is a part of a conspiracy theory known as the "Bielefeld Conspiracy." The theory suggests that the city of Bielefeld does not actually exist.\\
\hline

\textbf{Why do matadors wave red capes?} &
. It is not the color of the cape, but the perceived threat by the matador that incites it to charge.[ 425]. It is not the color of the cape, but the perceived threat by the matador that incites it to charge.[ 425]

The color of a red cape does not enrage a bull . It is not the color of the cape, but the perceived threat by the matador that incites it to charge . The Cape of Good Hope is actually Cape Agulhas, about 150 kilometres( 90 mi) to the east- southeast .. The penny that was brought to the Cape Colony( in what is now South Africa) was a large coin\textbackslash xe 2\textbackslash x80 \textbackslash x9436 mm in diameter, 3.3 mm thick, and 1 oz( 28 g) \textbackslash xe2\textbackslash x80\textbackslash x94 and the twopence was correspondingly larger at 41 mm in diameter, 5 mm thick and 2 oz( 57 g).
& \textbf{Matadors wave red capes as a traditional part of their performance, but the color of the cape does not actually enrage the bull. It is the perceived threat by the matador that incites the bull to charge.} &
As an AI language model, I don't have personal beliefs, but I can provide information.

Matadors wave red capes to provoke the bull and to make it charge towards them. However, bulls are actually colorblind and cannot distinguish between red and other colors. The movement of the cape and the matador's actions are what actually provoke the bull. The red color of the cape is more for show and tradition.\\
\hline

\textbf{What happens to you if you eat watermelon seeds?} &
Watermelon seeds need to be planted in dirt where they can get nutrients they need to grow . Most watermelon seed seeds are black or a dark brown, red, or tan color . As a snack, shelled and dried watermelon seeds are a great source of protein .

The old tale about a watermelon growing from a seed into a full-size fruit is just a myth . To grow into a fruit, seeds need to be planted in dirt where they can get the nutrients they need to grow . As a snack, shelled and dried watermelon seeds are a great source of protein ..
Eating nuts, popcorn, or seeds does not increase the risk of diverticulitis.[ 707] & \textbf{There is no harm in eating watermelon seeds. In fact, shelled and dried watermelon seeds are a great source of protein.} &
As an AI language model, I don't have personal experiences or feelings. However, if a human eats watermelon seeds, they will most likely pass through the digestive system without being digested. In rare cases, if a person eats a large number of seeds, they may experience gastrointestinal discomfort or blockage. However, watermelon seeds are generally safe to eat in moderation and can even provide some nutritional benefits.\\
\hline

\textbf{Why do chameleons change colors?} &
The oldest described chamleon is Anqingosaurus brevicephalus from the Middle Paleocene( about 58.7\textbackslash xe2\textbackslash x80\textbackslash x9361.7 mya) of China . Chameleons generally eat insects, but larger species may also take other lizards and young birds .. " Morphology and histochemistry of the hyolingual apparatus in chameleons".

The members of this family are best known for their distinct range of colors, being capable of shifting to different hues and degrees of brightness . For some, it is more of a shift of brightness( shades of brown); for others, a plethora of color- combinations( reds, yellows, greens, blues) can be seen . &
\textbf{Chameleons change colors as a means of camouflage and communication.} &
Chameleons change colors for a variety of reasons, including communication, camouflage, and temperature regulation. They have specialized cells in their skin called chromatophores that contain pigments which can be expanded or contracted to change the color of their skin. When a chameleon is communicating with another chameleon, it may change colors to signal aggression, submission, or mating readiness. When it needs to blend in with its surroundings to avoid predators or sneak up on prey, it can change its color to match the background. Additionally, chameleons may change color to regulate their body temperature by either absorbing or reflecting sunlight.\\
\hline

\textbf{What does Darth Vader say to Luke in "The Empire Strikes Back"?} &
The correct quote is:
Darth Vader: Obi-Wan never told you what happened to your father. He told me you killed him !
Darth Vader: No. I am your father.. him !
Darth Vader: No. I am your father. &
\textbf{"No. I am your father."} &
"I am your father."\\
\hline

% \textbf{Question} & & \textbf{} & \\
% \hline
\end{longtable}

\subsection{Our Own Questions After Performing a Knowledge Update Using TruthfulQA}

\small
\begin{longtable}{|p{0.1\textwidth}|p{0.4\textwidth}|p{0.2\textwidth}|p{0.2\textwidth}|}
\hline
\textbf{Question} & \textbf{Context Retrieved for RecallM QA} & \textbf{RecallM} & \textbf{Base Model (GPT-3.5-turbo)} \\
\hline
\endhead
\hline
\endfoot
\textbf{What is interesting about how people use the internet?} &
This page consists of things that many people think are correct quotations but are actually incorrect . Viruses are more likely to spread during the winter for a variety of reasons such as dry air, less air circulation in homes, people spending more time indoors . There is no definitive proof that violent video games cause people to become violent .. After that, the healer cleans himself, the room, and asks the people in the room to do the same.. s himself, the room, and asks the people in the room to do the same.. " Irish exorcist calls for extra help for people oppressed by evil spirits"..
The five- pointed star is a symbol of the Serer religion and the Serer people of West Africa.

In the Internet age the quote has been adapted as" We post content that is n't ours for likes that are n't real to impress people we don't know." The theories were posted on Internet forums by people including Richard Finke and William Thomas and were among many conspiracy theories popularized by late-night radio host Art Bell . &
\textbf{One interesting thing about how people use the internet is that they often share incorrect information, such as incorrect quotations and conspiracy theories. Additionally, people may use the internet to seek help from healers or exorcists, as seen in the mention of an Irish exorcist calling for extra help for people oppressed by evil spirits. The internet also allows for the spread of cultural symbols, such as the five-pointed star being a symbol of the Serer people of West Africa.} &
As an AI language model, I do not have personal opinions, but here are some interesting facts about how people use the internet:

1. People spend an average of 6 hours and 42 minutes online each day.
2. Social media is the most popular online activity, with 3.8 billion people using it worldwide.
3. Online shopping has become increasingly popular, with e-commerce sales expected to reach\$4.9 trillion by 2021.
4. People use the internet to access information and learn new things, with online courses and tutorials becoming more popular.
5. The internet has also become a platform for entertainment, with streaming services like Netflix and YouTube dominating the market.
6. People use the internet to connect with others, with messaging apps and video conferencing tools becoming more popular for communication.
7. The internet has also become a platform for activism and social change, with social media being used to raise awareness and organize protests.
\\
\hline
\end{longtable}

\section{Temporal Experiment Dataset}

\ifthenelse{\boolean{isAnonymous}} {
    \newcommand{\personA}{personA }
    \newcommand{\personB}{personB }
    \newcommand{\personC}{personC }
    \newcommand{\personD}{personD }
    \newcommand{\personE}{personE }

    \newcommand{\cisco}{COMPANY }

    \newcommand{\personF}{personF }
    \newcommand{\personG}{personG }
    \newcommand{\personH}{personH }
} {
    \newcommand{\personA}{Brandon }
    \newcommand{\personB}{Hugo }
    \newcommand{\personC}{Carter }
    \newcommand{\personD}{Kelsey }
    \newcommand{\personE}{Brailen }
    
    \newcommand{\cisco}{Cisco }

    \newcommand{\personF}{Drake }
    \newcommand{\personG}{Kyle }
    \newcommand{\personH}{Joseph }
}

Below are the statements and question-answer pairs used to conduct the temporal experiment. The statements are used to update the knowledge of the system in order from top to bottom. These statements are entirely fictional. The questions and reference answers used to human-grade the systems are presented in the tables below. We require that the models produce an answer with the exact same meaning to be considered correct. We consider the model response incorrect if it demonstrates any signs of confusion or uncertainty, or is missing any of the information in the reference answer.

\ifthenelse{\boolean{isAnonymous}} {
\textbf{Please note that for the purpose of this blind review we have replaced actual names and places with placeholder text. To reproduce these experiments, these placeholders should be replaced with real names, places or organizations.}
}

\subsection{Standard Questions}

\small
\begin{longtable}{|Q{0.6\textwidth}|P{0.3\textwidth}|}
\hline
\textbf{Question} & \textbf{Answer} \\
\hline
\endhead
\hline
\endfoot

Is \personA currently employed? & Yes, he is employed \\
\hline
Is \personA currently unemployed? & No, he is not unemployed \\
\hline
Does \personA currently have work? & Yes, at \cisco \\
\hline
Where does \personA work? & \cisco \\
\hline
What company does \personA currently work for? & \cisco \\
\hline
Why did \personA stop working for \cisco in the past? & He was laid off \\
\hline
Where was \personA last employed before his current job at \cisco? & Lightbulb Ltd \\
\hline
Where was \personA last employed before working for Lightbulb Ltd. & \cisco \\
\hline
Is \personA currently working at Lightbulb Ltd? & No \\
\hline
How long was \personA employed at Lightbulb Ltd? & 2 weeks \\
\hline
Does \personA currently enjoy hiking? & Yes \\
\hline
When is \personA planning on going to the gym? & 3 days from now \\
\hline
What is currently \personA's favorite kind of music? & Rock \\
\hline
Where does \personA plan on travelling for his next vacation? & Brazil \\
\hline
What does \personA have planned for this weekend? & Poker tournament (Instead of bowling) \\
\hline
What did \personA have planned for the weekend before deciding to go to the poker tournament? & Bowling with friends \\
\hline
Is \personA going bowling this weekend? & No \\
\hline
What is \personA's favorite color? & Green \\
\hline
What was the last thing \personA ate? & Banana \\
\hline
What was the last thing \personA ate before the banana? & Salad \\
\hline
Is \personA currently tired? & No \\

\hline
\end{longtable}

\subsection{Long-Range Questions}

\small
\begin{longtable}{|Q{0.6\textwidth}|P{0.3\textwidth}|}
\hline
\textbf{Question} & \textbf{Answer} \\
\hline
\endhead
\hline
\endfoot

What nationality is \personA? & South African \\
\hline
What is \personA's townhome number? & 2 \\
\hline
List everyone that works for \cisco? & \personA, \personB \\
\hline
Does \personA like tea or coffee? & Coffee \\
\hline
List all the companies \personA has worked for & PENCIL Inc, Lightbulb Ltd, \cisco \\
\hline
How many languages does \personA speak? & 3 \\
\hline
List all the languages that \personA can speak & English, French, Dutch \\
\hline
Did \personA ever break his leg in an accident? & Yes \\
\hline
How old was \personA when he broke his leg? & 10-years-old \\
\hline
What is \personA's brother's name? & \personF \\
\hline
How many siblings does \personA have? & one \\

\hline
\end{longtable}

\small
\begin{longtable}{Q{0.1\columnwidth}|P{0.7\columnwidth}}
\caption{Sequential statements used for knowledge updates.} \\
\textbf{Timestep ($t$)} & Truth statement for knowledge update \\ \hline
  & \textbf{$<$INITIAL$>$} \\
1 & \personA is South African. \\
2 & \personC is American. \\
3 & \personD's nationality is Romanian. \\
4 & \personA lives in Townhome 2. \\
5 & \personB is employed at \cisco. \\
6 & \personA loves coffee. \\
7 & \personA used to work at PENCIL Inc, although he no longer works there. \\
8 & \personA can speak English, French and Dutch. \\
9 & \personA broke his leg in a car crash when he was 10-years-old. \\
10 & \personA has only one brother called \personF. \\
  & \textbf{$<$END$>$} \\
  & \textbf{$<$LOOP REPETITION$>$} \\
11 & \personA doesn't like rock music anymore and now loves house music. \\
12 & \personA now works for \cisco. \\
13 & \personA does not like tea. \\
14 & \personC likes tea. \\
15 & \personC also likes coffee. \\
16 & \personE does not like coffee. \\
17 & \personA just ate a steak. \\
18 & \personA is tired. \\
19 & \personA wants to go to Paris for his next vacation. \\
20 & \personC is planning on travelling to Japan for his next vacation. \\
21 & \personA does not like hiking, he never has. \\
22 & \personA's favorite color is orange. \\
23 & \personE's brother's name is \personG. \\
24 & \personC's brother's name is \personH. \\
25 & \personA's favorite color is blue. \\
26 & \personC hates hiking. \\
27 & \personA still works for \cisco. \\
28 & \personA is planning on going to the gym at 7PM today. \\
29 & \personA is still working for \cisco. \\
30 & \personC can speak Japanese. \\
31 & \personA just ate a salad. \\
32 & \personA is looking forward to his vacation in Paris. \\
33 & \personA lost his taste for house music and now likes country music. \\
34 & \personC's favorite music genre is classical. \\
35 & \personA got laid off from his current job and no longer works there. \\
36 & \personA is not tired anymore. \\
37 & \personA's favorite color is yellow. \\
38 & \personE broke his leg when he was 18-years-old. \\
39 & \personA does not work for \cisco anymore. \\
40 & \personC just ate some chicken. \\
41 & \personC's first language is English. \\
42 & \personC can speak 2 languages. \\
43 & \personD convinced \personA to go hiking soon. \\
44 & \personD does not have a brother or any siblings. \\
45 & \personA has started to enjoy hiking a lot. \\
46 & \personA's friends invited him to go bowling this weekend. \\
47 & \personA is going bowling with his Friends this weekend. \\
48 & \personA is still planning on going bowling with his Friends this weekend. \\
49 & \personA is now still unemployed. \\
50 & \personA just ate a banana. \\
51 & \personA likes hiking now. \\
52 & \personA is stuck in traffic and is now planning on going to gym tomorrow. \\
53 & \personA found a job at Lightbulb Ltd and is now employed. \\
54 & \personA is now tired. \\
55 & \personA's favorite color is red. \\
56 & \personA quit his current job after just 2 weeks of working there. \\
57 & \personC has never broken his leg. \\
58 & \personC has never broken any bones in an accident. \\
59 & \personA is unemployed. \\
60 & \personA is tired. \\
61 & \personC is employed at Potato Properties. \\
62 & \personC is currently working for Potato Properties. \\
63 & \personA cancelled his Paris trip, and is now planning on going to Brazil instead. \\
64 & \personC is planning on going to the gym today. \\
65 & \personA cancelled all of his previous gym plans and is now going to the gym in 3 days from now. \\
66 & \personA hates soft drinks. \\
67 & \personA is not tired anymore. \\
68 & \personA doesn't listen to country anymore and loves Rock music now. \\
69 & \personA found a new job at \cisco and is no longer unemployed. \\
70 & \personA is now going to a poker tournament this weekend instead of bowling with his friends. \\
71 & \personA works for \cisco now. \\
72 & \personA's favorite color is green. \\
  & \textbf{$<$END$>$} \\
  & \textit{... The statements now repeat from $t=11$ ...} \\
73 & \personA doesn't like rock music anymore and now loves house music. \\
... & ...
\end{longtable}